\documentclass[letterpaper]{article} % DO NOT CHANGE THIS
\usepackage{aaai25}  % DO NOT CHANGE THIS
\usepackage{times}  % DO NOT CHANGE THIS
\usepackage{helvet}  % DO NOT CHANGE THIS
\usepackage{courier}  % DO NOT CHANGE THIS
\usepackage[hyphens]{url}  % DO NOT CHANGE THIS
\usepackage{graphicx} % DO NOT CHANGE THIS
\usepackage{natbib}  % DO NOT CHANGE THIS AND DO NOT ADD ANY OPTIONS TO IT
\usepackage{caption} % DO NOT CHANGE THIS AND DO NOT ADD ANY OPTIONS TO IT
\usepackage{algorithm}
\usepackage{algorithmic}
\usepackage{amsmath}
\usepackage{amssymb}
\usepackage{newfloat}
\usepackage{listings}
\DeclareCaptionStyle{ruled}{labelfont=normalfont,labelsep=colon,strut=off} % DO NOT CHANGE THIS
\floatstyle{ruled}
\newfloat{listing}{tb}{lst}{}
\floatname{listing}{Listing}
\title{TS-OOD: Evaluating Time-Series Out-of-Distribution Detection and Prospective Directions for Progress}
\author{
    Onat Gungor\textsuperscript{\rm 1},
    Amanda Rios\textsuperscript{\rm 2},
    Nilesh Ahuja\textsuperscript{\rm 2},
    Tajana Rosing\textsuperscript{\rm 1}
}
\affiliations{
    \textsuperscript{\rm 1}University of California, San Diego\\
    \textsuperscript{\rm 2}Intel Labs\\
    \{ogungor, tajana\}@ucsd.edu, \{amanda.rios, nilesh.ahuja\}@intel.com
}

\usepackage{bibentry}
\begin{document}

\maketitle

\begin{abstract}
Detecting out-of-distribution (OOD) data is a fundamental challenge in the deployment of machine learning models. From a security standpoint, this is particularly important because OOD test data can result in misleadingly confident yet erroneous predictions, which undermine the reliability of the deployed model. Although numerous models for OOD detection have been developed in computer vision and language, their adaptability to the time-series data domain remains limited and under-explored. Yet, time-series data is ubiquitous across manufacturing and security applications for which OOD is essential. This paper seeks to address this research gap by conducting a comprehensive analysis of modality-agnostic OOD detection algorithms. We evaluate over several multivariate time-series datasets, deep learning architectures, time-series specific data augmentations, and loss functions. Our results demonstrate that: 1) the majority of state-of-the-art OOD methods exhibit limited performance on time-series data, and 2) OOD methods based on deep feature modeling may offer greater advantages for time-series OOD detection, highlighting a promising direction for future time-series OOD detection algorithm development.  
\end{abstract}


\section{Introduction}
Safe and reliable deployment of machine learning (ML) models in real-world scenarios requires a classifier that not only ensures accurate classification of in-distribution (ID) samples but also reliably detects out-of-distribution (OOD) inputs, classifying them as unknown. This task is defined as OOD detection, which seeks to identify inputs that exhibit deviations from the distribution of the training data \cite{liang2017enhancing}. From a security standpoint, effective detection of OOD data is paramount for the deployment of ML models. OOD test data can lead to confidently erroneous predictions, thereby compromising the model's reliability and potentially introducing significant vulnerabilities. OOD data can be of a diverse nature and are often divided into: (1) ``Covariate Shift'' - OOD data which contains the same classes as in-distribution data (train) but presents significant changes in noise, acquisition parameters or other factors that will confuse the deployed model and may require specific adaptation procedures; (2) ``Semantic Shift'' - OOD data that contains unseen and completely novel unlearned classes w.r.t the in-distribution (train) data used by the deployed model. This category is noticeably harder to detect than the former \cite{sun2022dice,xu2023scaling}. 
% By addressing label shift, the model's robustness is enhanced to better handle dynamic and uncertain environments. 
% A key challenge in this context is label shift, where the distribution of labels in the test data may differ from that in the training data, often involving unseen classes that were not present during training. By addressing label shift, the model's robustness is enhanced to better handle dynamic and uncertain environments. 

Although a variety of models have been developed for OOD detection within the field of computer vision or language, such as ReACT \cite{sun2021react}, DICE \cite{sun2022dice}, and SCALE \cite{xu2023scaling}, the adaptation and application of these techniques to the time-series (TS) domain remain relatively limited and insufficiently explored. Yet, time-series data is widespread among industrial and security applications which require safe and robust deployed models. For instance, a sudden mechanical failure in a manufacturing plant could produce sensor readings corresponding to failure modes not represented in the training data, thus emphasizing the need for time-series OOD detection. Furthermore, time-series OOD detection is especially more challenging than vision due to several factors, among them: (1) the lack of foundational pre-trained time-series deep learning models; (2) a diverse array of data modalities with little in common among them except their variability across the time dimension, e.g. EEG vs. financial markets; (3) complex temporal relations across time-steps at varying seasonality frequencies; (4) many other TS specific characteristics, such as sparse peaks and fast oscillations \cite{belkhouja2023out} that vary across TS data modalities. In all, there is a critical need to advance research in time-series OOD detection to develop models that are more reliable, secure, and capable of effectively handling unexpected and novel data. 

%%%%%%%%%%%%%%%%%%%%%%%%%%%%%%%%%%%%%%%%%%%%%%%%%%%%%%%%%%%%%%%%%%%
\begin{figure*}[t]
    \centering
    \captionsetup{justification=centering}
    \includegraphics[width=14cm]{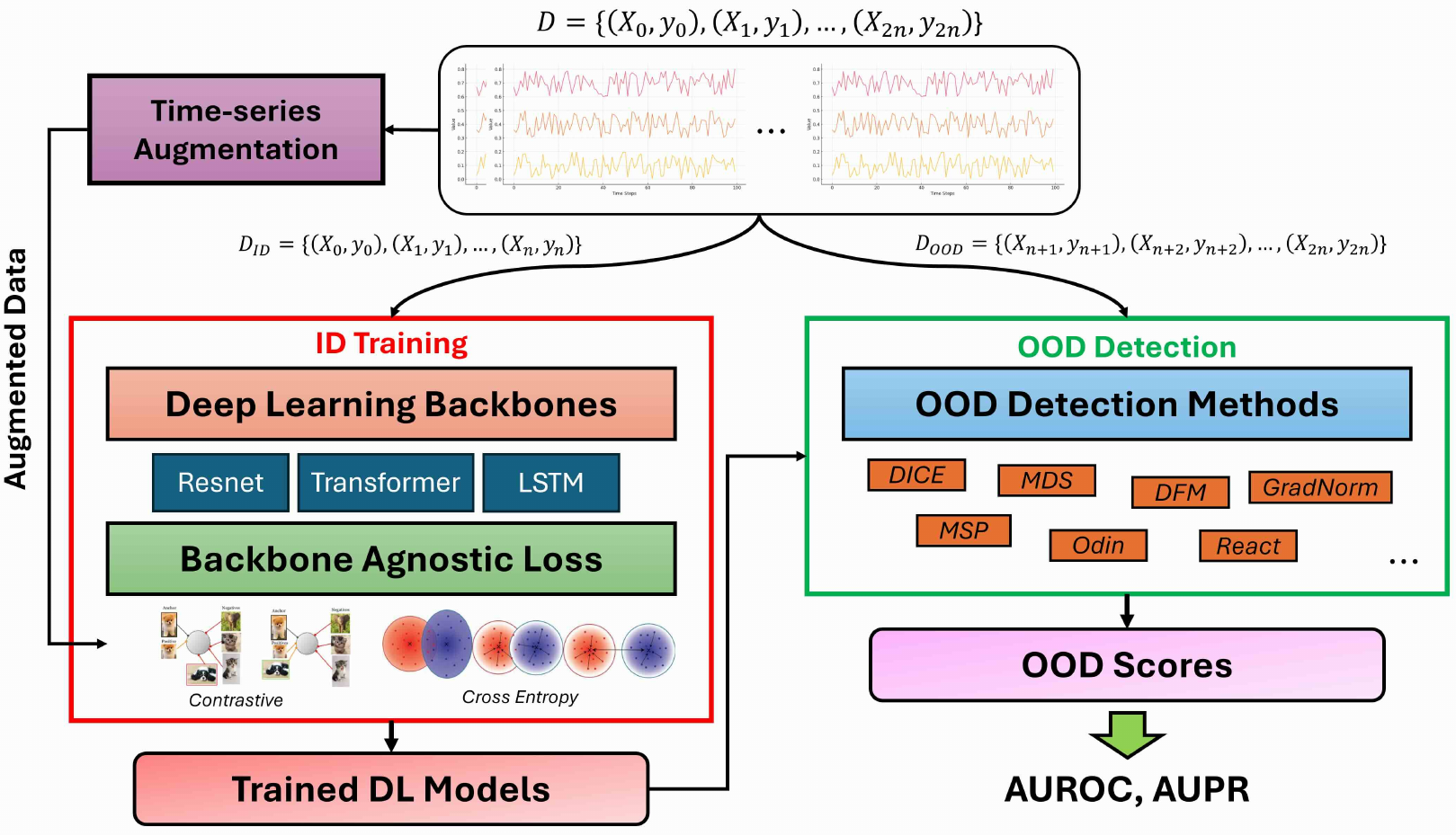}
    \caption{The proposed evaluation pipeline (TS-OOD) for modality-agnostic OOD detection models across a range of deep learning architectures, time-series data augmentations, and loss functions on multivariate time-series data}
    \label{fig:TS-OOD-framework}
\end{figure*}
%%%%%%%%%%%%%%%%%%%%%%%%%%%%%%%%%%%%%%%%%%%%%%%%%%%%%%%%%%%%%%%%%%%

\textbf{Related Works:} Surprisingly, few studies analyze and/or propose OOD models specifically designed for TS OOD. Among these few, seasonal ratio scoring (SRS) \cite{belkhouja2023out} uses seasonal patterns to calculate a score based on seasonal and residual component ratios. The primary limitations of this study are the computationally expensive STL decomposition step and the requirement for an additional deep learning model (conditional variational autoencoder) for OOD detection. These factors raise concerns about the scalability and efficiency of the proposed solution. 
% It Trains two Conditional Variational Autoencoders (CVAE) to capture the seasonal and residual components, which are then used to compute the SRS. If the score of a new data point falls within a learned threshold range, it's considered in-distribution; otherwise, it’s flagged as OOD. 
Another example, Diversify \cite{lu2024diversify} employs a unique diversification strategy that enhances OOD detection and the generalization of time-series models via adversarial training. 
% The method uses a min-max adversarial approach, where it first segments the time-series data into various latent sub-domains by maximizing the distribution differences between segments. Concurrently, it aims to learn representations that are invariant across these latent domains by minimizing the distribution divergence among them. The main limitations of this study include the indeterminate number of latent distributions a priori and the inherent instability in training associated with adversarial-based method. 
Additionally, Codit \cite{kaur2023codit} employs conformal prediction, leveraging deviations from in-distribution temporal equivariance as the non-conformity measure, to measure the reliability of post-deployment predictions. 
While these time-series specific works provide an important advancement to the TS OOD field, they share some major limitations. Firstly, they offer very limited comparisons to existing state-of-the-art modality agnostic OOD methods. Without such comparisons, it is not possible to accurately assess the impact of time-series specific algorithmic additions (e.g., seasonality decomposition, temporal equivariance) on overall OOD detection performance. Secondly, these studies evaluate semantic OOD detection by introducing unknown novel classes from entirely different datasets/data-modalities. In the case of multivariate time-series, this can be a much easier task due to the significant diversity in the origins of time-series data. For instance, classes derived from EEG sensors can differ drastically from those associated with ECG sensors or industrial sensor networks, making OOD detection relatively straightforward. In both vision and language modalities, it has become a consensus that evaluating semantic OOD detection between orthogonal sets of classes from the same original dataset—ensuring consistency in time-series origin, number of channels, and other relevant statistical factors—presents a more challenging and realistic experimental scenario.  
% OOD between two different datasets is known to be much easier then between different classes of same dataset (same low level acquisition stats). Cite many vision OOD papers that have shown this
% So our work differentiates itself, for time-series, mainly in this regard.
% However, the state-of-the-art OOD detectors and datasets employed are domain-specific in this study.

\textbf{Our Contributions:} In this paper, we endeavor to fill some of the aforementioned gaps in TS OOD detection by conducting a comprehensive evaluation of semantic time-series OOD detection. We provide an in-depth analysis of modality-agnostic OOD model detection performance across a range of deep learning architectures, TS data augmentations, and loss functions. Moreover, and very importantly, we evaluate all models under a more challenging and realistic semantic OOD evaluation setting with ID and OOD data sets created from the same TS dataset in order to avoid low-level statistics differentiation between ID and OOD sets. Overall, our results indicate that the majority of modality-agnostic OOD detection methods \cite{hendrycks2016baseline,liu2020energy,sun2021react,sun2022dice} do not yield reliable measurements for time-series data. We suggest that OOD detection approaches leveraging deep feature modeling \cite{ahuja2019probabilistic} may be more effective for time-series data, highlighting a promising direction for TS OOD algorithms that can be developed based on this technique. 

\section{TS-OOD: Evaluation Pipeline}
Figure \ref{fig:TS-OOD-framework} illustrates our time-series out-of-distribution (TS-OOD) evaluation pipeline. TS-OOD initially generates in-distribution (ID) and out-of-distribution (OOD) datasets derived from the same time-series dataset. To obtain trained deep learning (DL) models, TS-OOD employs multiple DL architectures, utilizing backbone-agnostic loss functions to train on ID data. During out-of-distribution detection, a diverse set of state-of-the-art methods is employed to compute OOD scores and evaluate final performance metrics utilizing the trained ID model. 

\subsection{OOD detection setting}
\label{sec:ood-se}
In our experiments, the test data is an unsupervised mixture of unseen in-distribution (ID) samples ("known" classes) and out-of-distribution (OOD) samples ("new" unseen and unlearned classes),
% \[
% \text{ID}_t = \{ u_{\text{old}}, u_{\text{unseen}} \mid u_{\text{old}}, u_{\text{unseen}} \sim D_k \}, \quad k = 1, \dots, t-1
% \]
% \[
% \text{OOD}_t = \{ u_{\text{new}}, u_{\text{unseen}} \mid u_{\text{new}}, u_{\text{unseen}} \sim D_t \}
% \]
where ID data consists of unseen samples of known classes, sampled from the same distribution used in training of the deployed classification model. In contrast, OOD data contains samples from an entirely new distribution comprised of unseen and entirely unknown novel classes. The goal of the OOD detector is to accurately differentiate between ID and OOD. Given the number of classes \( C \) in a dataset, we designate the first half as in-distribution (ID) and the second half as out-of-distribution (OOD). For training the deep learning (DL) models (classifiers), we use only the ID portion of the training data. 
% For OOD detection, the test data includes both ID and OOD samples, with the goal of detecting the OOD samples.

\subsection{In-distribution (ID) Training}
\subsubsection{Deep Learning (DL) Backbones.} We select three commonly used time-series DL backbones to assess their performance in OOD detection. 
\begin{itemize}
    \item \textbf{Residual Network (ResNet) \cite{wang2017time}.} We adapt ResNet by using 1-D convolutional layers. Our ResNet consists of three residual blocks, each containing three convolutional layers, followed by a global average pooling layer and a softmax layer.
    \item \textbf{Transformer (TST) \cite{vaswani2017attention}.} TST utilizes self-attention mechanisms to process and model long-range dependencies in sequential data, and they are increasingly used in time-series analysis for their ability to capture complex temporal patterns and relationships. We stack three TST encoder layers with a classification head.
    \item \textbf{Long Short-term Memory (LSTM) \cite{hua2019deep}.} LSTM learns long-term dependencies in sequential data by using memory cells and gates, and it is commonly used in time-series analysis due to its ability to model temporal dependencies and patterns. We stack two LSTM layers, connected by a fully connected layer.  
\end{itemize}
\subsubsection{Time-series Data Augmentation.}
Time-series data augmentation methods are a fundamental component of contrastive learning algorithms. Contrastive learning has demonstrated tremendous success in vision and language, owing to its ability to generate more generalizable data embeddings (features) compared to traditional loss functions such as Cross-Entropy \cite{liu2024guidelines}. Behind their success lies the enforcement, through carefully constructed augmentations, of key modality-specific invariances. For instance, in vision, augmentations that alter object positioning, viewpoint, and texture enforce positional, viewpoint, and textural invariance in the learned object representations within the deep learning embedding. A major difficulty in proposing and selecting augmentations for time-series data is that invariances are less intuitive due to the sheer diversity of time-series data origins, ranging from medical to financial domains. The lack of guidance regarding which invariances should be enforced, and whether they are modality-specific to time-series data, presents significant challenges to deep learning in time-series and remains an active area of research. Because the performance of OOD detection methods is directly influenced by the choice and application of augmentations in the backbone, we apply and compare a diverse set of state-of-the-art time-series augmentations to assess their impact on OOD detection. 
\begin{itemize}
    \item \textbf{Jittering} modifies a time series dataset by adding random noise, such as Gaussian, Poisson, or Exponential \cite{sarkar2021detection}.
    \item \textbf{Permutation} consists of segmentation and permuting. While segmentation divides the series into multiple subsequences, permuting randomly rearranges the subsequences \cite{jiang2021self}.
    \item \textbf{Magnitude Warping} changes the magnitude of each sample in a TS dataset by multiplication of the original time series with a cubic spline curve \cite{um2017data}.
    \item \textbf{Window Warping} either speeds up or down the selected windows in TS data. Then, the whole resulting time series is scaled back to the original size in order to keep the timesteps at the original length \cite{rashid2019window}.
    \item \textbf{Resizing} consists of cropping and resizing. While cropping keeps only a fraction of an original time series segment, resizing adjusts the cropped segment back to the original length \cite{chen2021clecg}.
    \item \textbf{Flipping} reverses the time steps in the sequence, altering the data’s chronological order \cite{sarkar2021detection}.
    \item \textbf{Time Masking} drops out certain observations within the time series to augment the data \cite{han2021semi}. 
\end{itemize} 
\subsubsection{Backbone Agnostic Loss Functions.} We employ two different backbone-agnostic loss functions to assess their impact on time-series OOD detection performance.
\begin{itemize}
    \item \textbf{Cross-entropy (CE) loss \cite{mao2023cross}.} CE loss is commonly employed in time series classification tasks, particularly when the goal is to assign categorical labels to temporal data, as it effectively measures the discrepancy between the predicted probability distributions and the true class labels across time-dependent sequences. CE is formulated as: 
    \begin{equation}
    \mathcal{L} = - \sum_{i=1}^{C} y_i \log(p_i)
    \label{eq:CE-loss}
    \end{equation}
    where \( C \) is the number of classes, \( y_i \) is the true label, and \( p_i \) is the predicted probability for class \( i \).
    \item \textbf{Multi-positive contrastive (MPC) loss  \cite{tian2024stablerep}.} The MPC loss is a contrastive learning-based adaptation of the cross-entropy loss, where the interpretations of \( y_i \) and \( p_i \) change. Consider an encoded anchor sample \( a \), and a set of encoded candidates \( \{b_1, b_2, \dots, b_K\} \). We first compute the probability \( p \) that describes how likely \( a \) is to match each \( b \):
    \begin{equation}
    p_i = \frac{\exp(a \cdot b_i / \tau)}{\sum_{j=1}^{K} \exp(a \cdot b_j / \tau)}
    \end{equation}
    where \( \tau \in \mathbb{R}^+ \) is the scalar temperature hyperparameter, and \( a \) and all \( b \) have been \( \ell_2 \) normalized. This represents a \( K \)-way softmax classification distribution applied to all encoded candidates. Assume there is at least one candidate that the anchor \( a \) matches. Then, the ground-truth categorical distribution \( y \) is formulated as:
    \begin{equation}
    y_i = \frac{\mathbf{1}_{\text{match}(a, b_i)}}{\sum_{j=1}^{K} \mathbf{1}_{\text{match}(a, b_j)}}
    \end{equation}
    where the indicator function \( \mathbf{1}_{\text{match}(\cdot, \cdot)} \) indicates whether the anchor and candidate match. Then, the MPC loss can be formulated as the CE loss, as shown in Equation \ref{eq:CE-loss}. 
    %It is important to note that the MPC loss is highly sensitive to the selection of positive augmentations, as elaborated in the preceding section.

\end{itemize}

\subsection{Out-of-distribution (OOD) Detection}
\subsubsection{OOD Detection Methods.} We select eight state-of-the-art post-hoc modality-agnostic OOD detection methods to evaluate their applicability to time-series OOD detection. 
% The selected methods do not require additional tuning to OOD data, neither do they necessarily need to be concurrently trained with the deployed main model. training or complex models, making them efficient for OOD detection. 
We include classification-based, distance-based, and density-based methods which represent some of the main categories of modality-agnostic OOD detection methods in the existing literature \cite{yang2024generalized}:
\begin{itemize}
    \item \textbf{Maximum Softmax Probability (MSP) \cite{hendrycks2016baseline}} detects OOD samples by calculating the maximum softmax probability (MSP) and classifying inputs with a low MSP as OOD. 
    %Variants of MSP, that also compute uncertainty from the classification layer logit distribution, such as Entropy and Margin [cite] provide similar results.
    \item \textbf{ODIN \cite{liang2017enhancing}} improves upon MSP by incorporating temperature scaling and input perturbation to enhance OOD detection performance. 
    \item \textbf{Energy-based OOD Detection (EBO) \cite{liu2020energy}} uses the model's energy function, derived from logits, to identify OOD samples. OOD samples result in high energy values, reflecting the model's uncertainty. 
    \item \textbf{GradNorm \cite{huang2021importance}} leverages the gradients of the model's output with respect to the input. The method detects OOD samples by measuring the gradient magnitude, where large or irregular gradients indicate data that differs from the training distribution. 
    \item \textbf{ReACT \cite{sun2021react}} detects OOD samples by analyzing rectified activations within the model. It measures how structured or unstructured the activations are, with weak activations signaling potential OOD data.
    \item \textbf{Dice \cite{sun2022dice}} measures the degree of sparsity in the activations to identify OOD samples. For OOD samples, the activations become sparse.
    \item \textbf{Mahalanobis (MDS) \cite{lee2018simple}} introduced the procedure of computing OOD scores using features extracted from several depths of a pre-trained deep neural network. For that, they use ID class conditioned Gaussians and a tied covariance to then compute the OOD score via the Mahalanobis distance of a sample to the modeled ID Gaussian distributions. Naturally, for OOD samples, this distance tends to be large. Note that for multivariate time-series, due to the absence of large-scale (foundation-like) DNNs, we train the backbone with the ID class samples only. 
    % between the input sample and the class distributions by utilizing the class mean and covariance extracted as embeddings from deep layers of a deep neural network. It measures how far the input is from the center of the in-distribution (ID) class. For OOD samples, this distance tends to be large.
    \item \textbf{Deep Feature Modeling (DFM)  \cite{ahuja2019probabilistic}} uses an equivalent procedure (to MDS) of computing OOD scores from several depths of a pre-trained deep network. Yet, they model the ID classes with a different approach. Their assumption is that features induced by the ID training dataset do not fully span the high dimensional space in which they reside in the DNN layers. As such, they propose to project the features of each ID class extracted from a given deep layer to a lower dimensional embeddings via Principle Component Analysis (PCA). The OOD score is then computed as the ``Feature Reconstruction Error'' of a sample projected to the low dimension and then re-projected back via an ID class' inverse PCA transform. For this paper, we extend the concept of modeling from DNN low or high level per-layer features as ``Deep Feature Modeling (DFM)'' irrespective of the ID modeling procedure, e.g. Gaussian, PCA (hereon referred to as DFM-PCA), or other statistical approach. As such, we introduce other versions of DFM-like OOD models using Isolation Forest (DFM-IF) and One-class Support Vector Machine (DFM-OCSVM) where per layer features are modeled, per ID class, using either IF or OCSVM. Similarly, a sample which is OOD will tend to have a large distance to an ID class trained IF, OCSVM or PCA reconstruction. Note that, similar to Mahalanobis we train the DNN backbone with ID classes' train data. We experimented with extracting features from several depths of the backbones tested and obtained on average superior results using the pre-logit layer. We leave the exploration of more complex feature extraction and modeling to future work. 
    % focuses on modeling the deep features of a DNN using a probabilistic distribution. The method detects OOD samples by measuring the likelihood of the deep features under the learned distribution, where low likelihoods indicate OOD inputs. To map the high-dimensional features onto an appropriate lower-dimensional subspace, they utilize Principal Component Analysis (PCA), Isolation Forest (IF), and One-class Support Vector Machine (OCSVM). These unsupervised algorithms learn a per-class mapping to a lower-dimensional subspace using the training data and compute the feature reconstruction error (FRE) on the test data. The underlying intuition is that OOD samples will fall outside the subspace defined by the ID samples, thereby yielding higher FRE scores. We present three types of DFMs: DFM-PCA, DFM-IF, and DFM-SVM.  
\end{itemize}

\subsubsection{OOD Scoring and Evaluation}
Each OOD detection method outputs scores for both ID and OOD test samples. After binarizing the true test labels (ID vs. OOD), we calculate the Area Under the Receiver Operating Characteristic curve (AUROC) and the Area Under the Precision-Recall curve (AUPR) to evaluate OOD detection performance. We select AUROC and AUPR for their threshold independence, allowing for a comprehensive evaluation across all decision thresholds.

\section{Experimental Setup}
We employ multivariate datasets from the UCR/UEA time series repository \cite{dau2019ucr}. Specifically, we select ArticularyWordRecognition (AWR), Epilepsy (EP), EthanolConcentration (EC), HandMovementDirection (HMD), Handwriting (HW), Libras (LIB), LSST, NATOPS (NATO), PEMS-SF (PEMS), PenDigits (PD), PhonemeSpectra (PS), RacketSports (RS), and UWaveGestureLibrary (UW). Table \ref{tbl:datasets} presents the selected multivariate time-series datasets along with their characteristics. For all datasets, we use half the classes as in-distribution and the other half as out-of-distribution. The in-distribution classes/data are used to train the backbones and OOD models. OOD models are not trained/tuned on OOD data, only evaluated on the latter. Also,  for each dataset the evaluation data will contain a 50/50 split of unseen samples from ID classes and unseen and unknown samples from the OOD classes.

%%%%%%%%%%%%%%%%%%%%%%%%%%%%%%%%%%%%%%%%%%%%%%%%%%%%%%  
\begin{table}[t]
\centering
\caption{Selected Datasets from the UEA Repository}
\scalebox{0.68}{
\begin{tabular}{cccccc}
\hline
\textbf{Dataset}  & \textbf{Train} & \textbf{Test} & \textbf{Dims} & \textbf{Length} & \textbf{Classes} \\ \hline
ArticularyWordRecognition (AWR)     & 275            & 300           & 9             & 144            & 25                \\
Epilepsy (EP)          & 137            & 138           & 3             & 206           & 4                \\
EthanolConcentration (EC)          & 261            & 263           & 3             & 1751           & 4                \\
HandMovementDirection (HMD)         & 160            & 74           & 10             & 400           & 4                \\
Handwriting (HW)          & 150            & 850           & 3             & 152           & 26                \\
Libras (LIB)       & 180            & 180           & 2             & 45           & 15                \\
LSST          & 2459            & 2466           & 6             & 36           & 14                \\
NATOPS (NATO)            & 180            & 180           & 24            & 51            & 6                \\
PEMS-SF (PEMS)            & 267            & 173           & 963            & 144            & 7                \\
PenDigits (PD)          & 7494            & 3498           & 2            & 8            & 10                \\
PhonemeSpectra (PS)            & 3315            & 3353           & 11            & 217            & 39                \\
RacketSports (RS)     & 151            & 152           & 6             & 30            & 4                \\
UWaveGestureLibrary (UW)   & 120           & 320          & 3             & 315           & 8                \\ \hline
\end{tabular}
}
\label{tbl:datasets}
\end{table}
%%%%%%%%%%%%%%%%%%%%%%%%%%%%%%%%%%%%%%%%%%%%%%%%%%%%%%

% \subsection{Time-series Data Augmentation}
% We use Time Series Generative Modeling (TSGM) \cite{nikitin2023tsgm} and tsaug \cite{tsaug} libraries to create the time-series augmented data.   

% \subsection{DL Model Training}
% Optimizer, learning rate, etc. 
% Mention about the training process of CL-based loss

% \subsection{OOD Detection Methods}
% We utilize the publicly available repositories for each out-of OOD detection method, employing the default parameters specified for each method. 

% \subsection{Hardware} 
% We conducted our experiments on a Linux virtual machine server equipped with a 32-core CPU, 64 GB of RAM, and an NVIDIA RTX 2080 Ti GPU.

%%%%%%%%%%%%%%%%%%%%%%%%%%%%%%%%%%%%%%%%%%%%%%%%%%%%%
\begin{figure}[t]
    \centering
    \captionsetup{justification=centering}
    \includegraphics[width=1\linewidth]{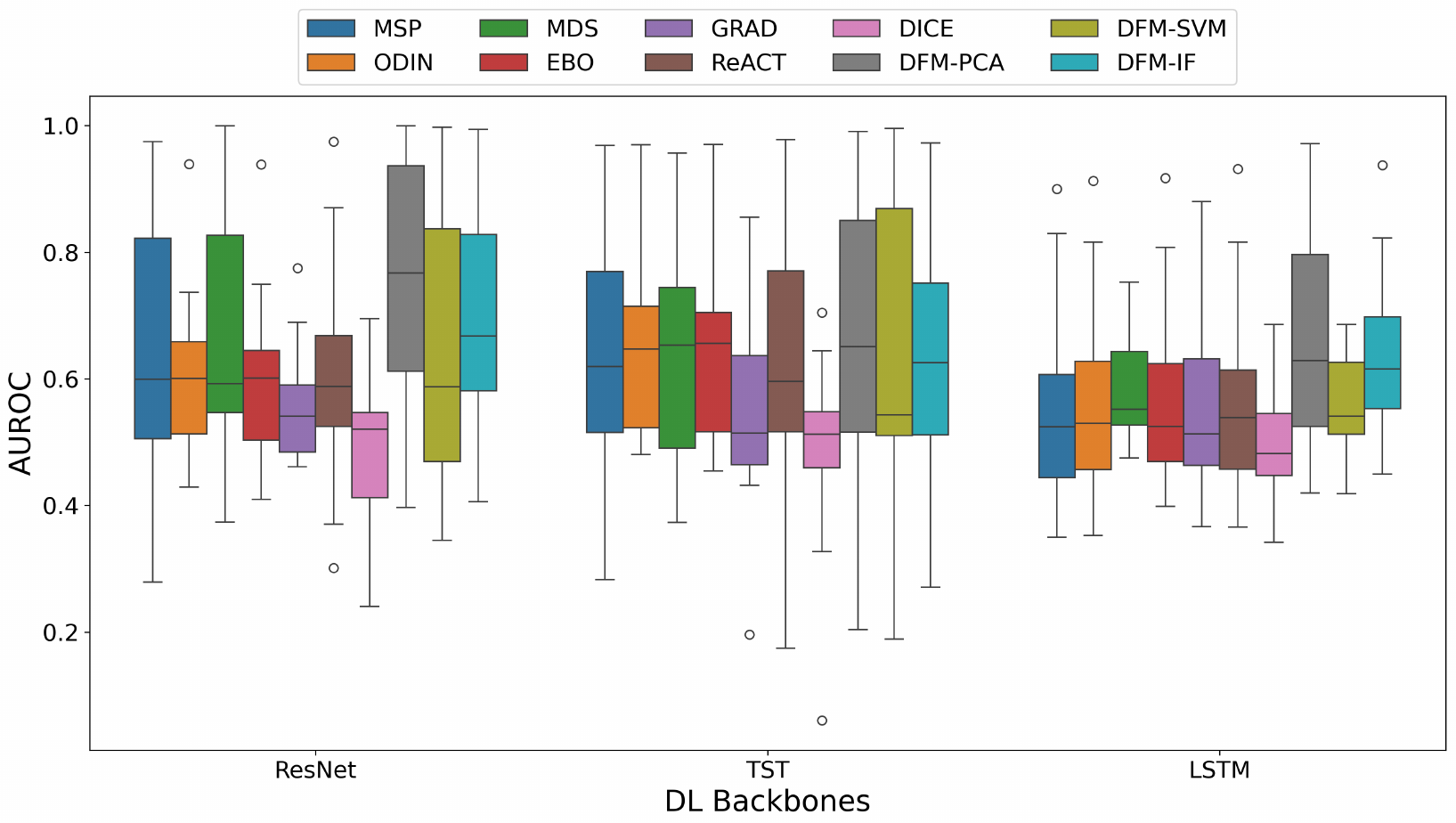}
    \includegraphics[width=1\linewidth]{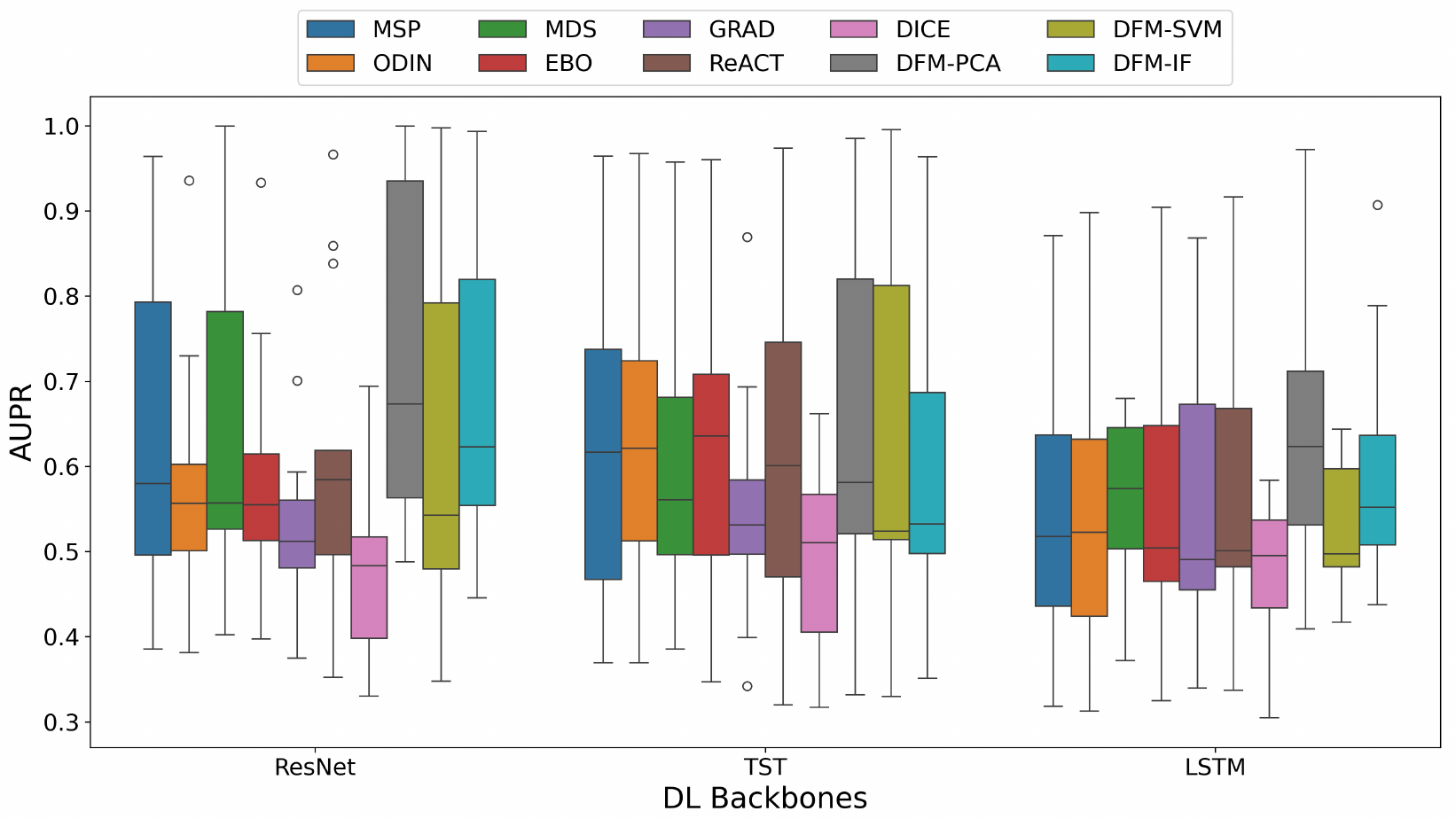}
    \caption{Impact of DL backbone on OOD detection performance across state-of-the-art deep-learning OOD detection methods}
    \label{fig:backbone-analysis}
\end{figure}
%%%%%%%%%%%%%%%%%%%%%%%%%%%%%%%%%%%%%%%%%%%%%%%%%%%%%

\section{Results}
\subsection{Sensitivity to DL Backbone Architecture}

Figure \ref{fig:backbone-analysis} presents the out-of-distribution (OOD) detection performance for each deep learning (DL) backbone. The top panel shows the AUROC scores, while the bottom panel illustrates the AUPR scores. Each OOD detection method is represented by a different color. While the x-axis denotes the DL backbone, the y-axis shows the corresponding score. We can observe that the performance of LSTM is significantly inferior to that of ResNet and TST. Although the performance of ResNet and TST is comparable, ResNet demonstrates better average OOD detection performance compared to TST.  It is evident that the majority of state-of-the-art OOD detection methods, e.g., DICE (pink), ReACT (brown), fail to produce reliable measurements for time-series data. Besides, it is important to highlight that DFM-based methods, such as DFM-PCA (gray) and DFM-SVM (light green), exhibit superior performance relative to other approaches throughout the tested backbones. For the remainder of the paper, we will use the ResNet architecture since it yielded the best averaged results across datasets and methods.

\subsection{Impact of Augmentation Techniques}
\begin{figure}[t]
    \centering
    \captionsetup{justification=centering}
    \includegraphics[width=1\linewidth]{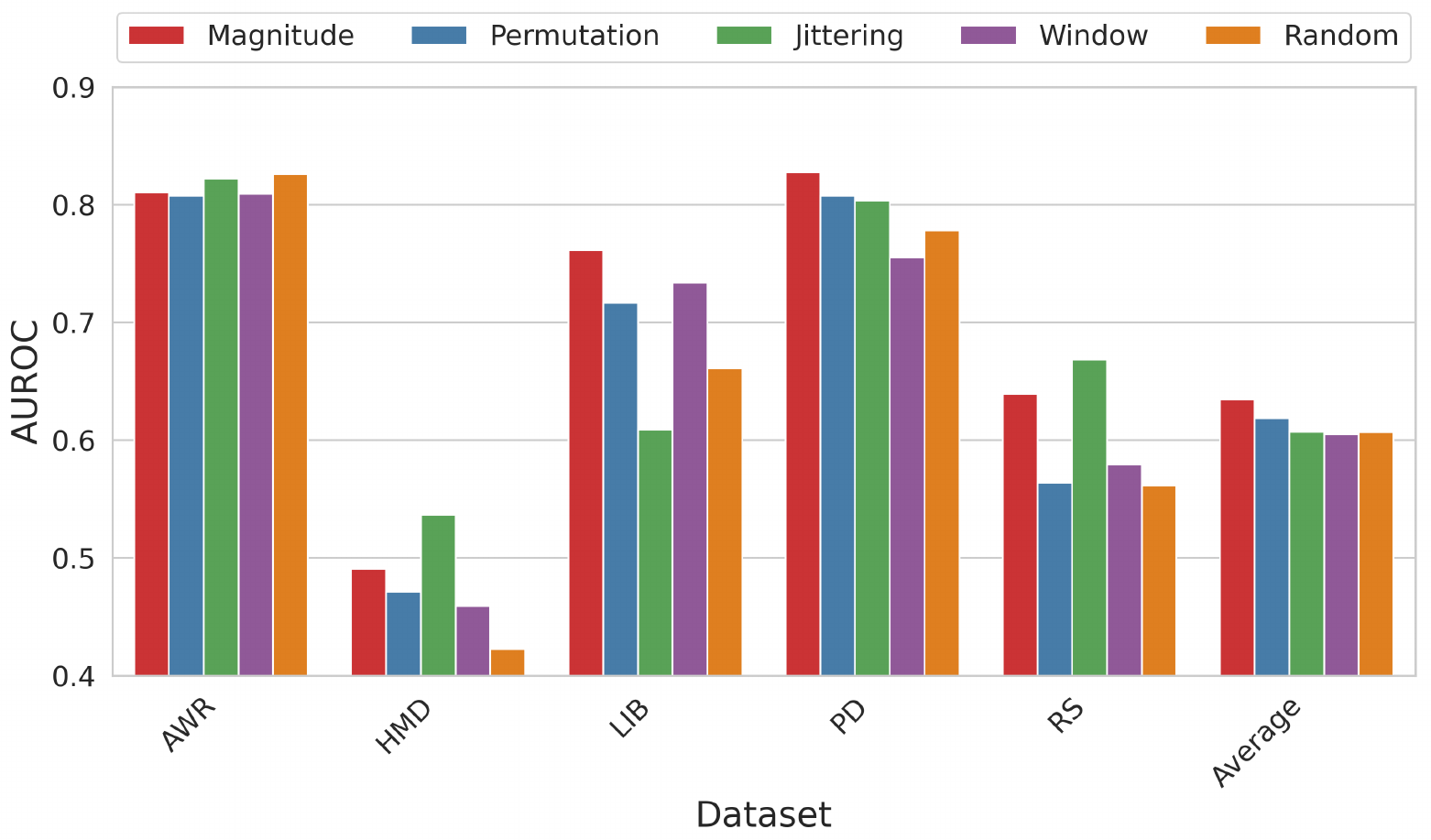}
    \caption{Augmentation Analysis on select datasets}
    \label{fig:aug-analysis}
\end{figure}

Data augmentation plays a pivotal role in contrastive learning by generating diverse representations of the same data, thereby facilitating the model's ability to learn robust, invariant features and improving its generalization capabilities \cite{tian2020makes}. Consequently, it is essential to critically evaluate the impact of data augmentation on OOD detection, as the efficacy of augmentation strategies may vary when addressing data that falls outside the distribution of the training set. 
Figure \ref{fig:aug-analysis} illustrates the results of our augmentation analysis on selected datasets (designated by their abbreviations introduced in Table 1), with the AUROC scores averaged across various OOD detection methods. We can clearly see that the selected augmentation impacts the OOD detection performance. For instance, at the LIB dataset, the AUROC score for the Magnitude augmentation method is 0.76, whereas the score decreases to 0.61 when Jittering is used as the augmentation method. Moreover, augmentation performance and ranking are very variable across datasets, which is expected since these datasets have very different TS origins. On average, the Magnitude augmentation method yields the highest performance, with an AUROC score of 0.634, followed by the Permutation method, which achieves a score of 0.619. Building on this result, we will present the contrastive learning outcomes for the Magnitude augmentation method in the subsequent subsections. Note that several combinations of the individual augmentations can be used during contrastive training, and we leave the analysis of a more exhaustive combination strategy to future research. 

%%%%%%%%%%%%%%%%%%%%%%%%%%%%%%%%%%%%%%%%%%%%%%%%%%%%%
%%%%%%%%%%%%%%%%%%%%%%%%%%%%%%%%%%%%%%%%%%%%%%%%%%%%%

\subsection{Selection of the Training Loss function on post-hoc OOD detection methods}
\begin{figure}[t]
    \centering
    \captionsetup{justification=centering}
    \includegraphics[width=1\linewidth]{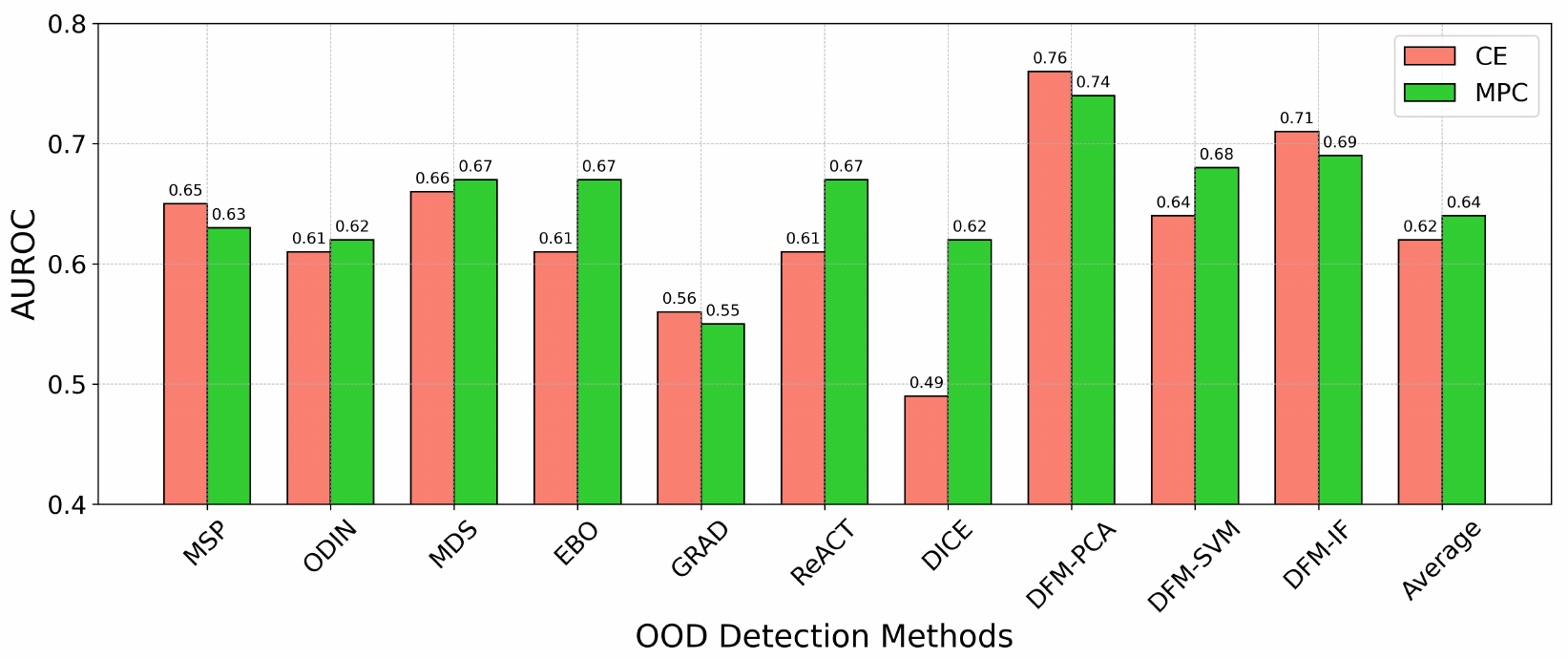}
    \includegraphics[width=1\linewidth]{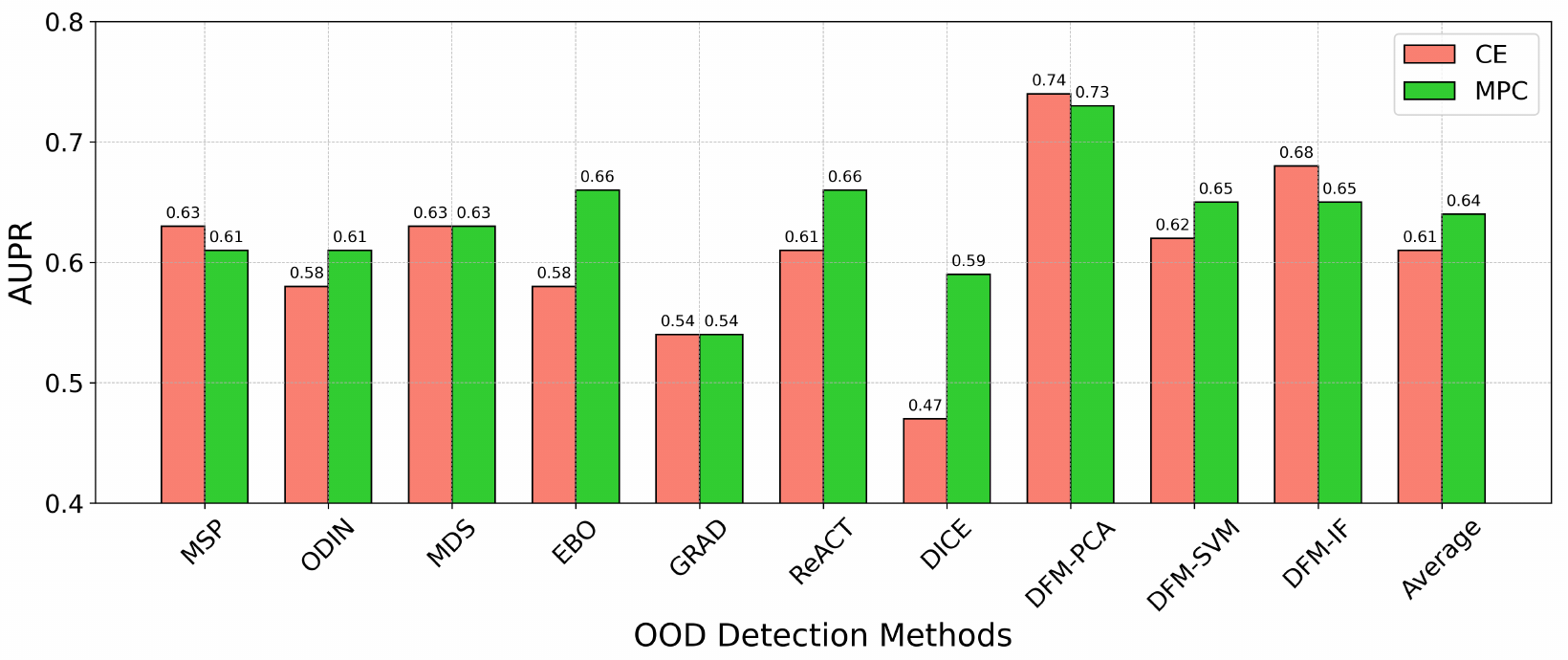}
    \caption{Loss Function Analysis}
    \label{fig:loss-analysis}
\end{figure}

To evaluate which loss function yields better OOD detection performance, we select the best augmentation method for the contrastive loss (MPC) and compare it with the cross-entropy loss (CE) on ResNet. This comparison is essential for identifying the most effective approach to enhancing the robustness and reliability of OOD detection in time-series data. Figure \ref{fig:loss-analysis} shows the AUROC (top) and AUPR (bottom) scores, averaged across the datasets, for each OOD detection method. The contrastive loss-based training appears to enhance the performance of CE-loss for certain OOD detection methods, such as ReACT and DICE. This improvement may be attributed to the similar approach employed by both methods, which involves utilizing activations. On average, MPC outperforms CE by 4.3\% in AUROC and 4.1\% in AUPR, with notable gains observed in methods such as ReACT, and DICE, where the contrastive loss-based training leads to more significant enhancements in OOD detection performance. However, certain methods, such as MSP and ODIN, exhibit minimal improvements, indicating that the effectiveness of MPC may depend on the specific characteristics and underlying mechanisms of the OOD detection method. Future work could focus on designing a TS-specific contrastive loss function, which may enhance OOD detection performance by more effectively capturing the unique temporal dependencies inherent in time-series data.

%%%%%%%%%%%%%%%%%%%%%%%%%%%%%%%%%%%%%%%%%%%%%%%%%%%%%
%%%%%%%%%%%%%%%%%%%%%%%%%%%%%%%%%%%%%%%%%%%%%%%%%%%%%

% \subsection{Determining Feature Extraction Points}
% Per layer analysis. 

%%%%%%%%%%%%%%%%%%%%%%%%%%%%%%%%%%%%%%%%%%
\begin{table}[t]
\centering
\captionsetup{justification=centering}
\begin{tabular}{|c|c|c|}
\hline
\textbf{OOD Method} & \textbf{ResNet-CE} & \textbf{ResNet-MPC} \\ \hline
\textbf{DFM}        & \textbf{0.885}              & 0.703               \\ \hline
\textbf{MDS}        & 0.690              & 0.751               \\ \hline
\textbf{EBO}        & 0.710              & \textbf{0.753}               \\ \hline
\textbf{MSP}        & 0.491              & 0.439               \\ \hline
\textbf{ReACT}      & 0.400              & 0.520               \\ \hline
\textbf{ID Acc}        & 0.737              & 0.721               \\ \hline
\end{tabular}
\caption{Pearson correlation coefficient (PCC) among ID accuracy vs OOD ROC-AUC}
\label{tbl:ood-correl}
\end{table}
%%%%%%%%%%%%%%%%%%%%%%%%%%%%%%%%%%%%%%%%%%

\subsection{ID vs OOD Performance Correlation}
Previously, it has been shown that out-of-distribution performance is strongly correlated with in-distribution performance across a wide range of models for vision tasks \cite{miller2021accuracy}. However, these results are based on limited visual shifts or artificial noise, raising questions about the generalizability of this correlation when severe shifts occur in OOD data, e.g. semantic shifts, such as in our experimental setting \cite{shi2024lca}. Here, we examine the ID vs. OOD correlation in time series data. Table \ref{tbl:ood-correl} presents the correlations between ID accuracy (based on ID test data) and OOD ROC-AUC scores for the selected OOD methods. We also report the in-distribution validation accuracy for the CE and MPC loss functions in the last row. We specifically selected these OOD methods because they represent different OOD approaches: distance-based (DFM, MDS), classification-based (MSP, ReACT), and density-based (EBO). We observe a strong ID-OOD correlation with DFM, MDS, and EBO, while classification-based OOD methods yield a weaker correlation. This suggests that distance-based and density estimation approaches may be more effective for time-series OOD detection.    

%%%%%%%%%%%%%%%%%%%%%%%%%%%%%%%%%%%%%%%%%%%%%%%%
\begin{table}[t]
\centering
\caption{Overhead (ms) of selected OOD methods}
\scalebox{0.9}{
\begin{tabular}{|l|c|c|c|c|c|c|}
\hline
\textbf{Method} & \textbf{MSP} & \textbf{MDS} & \textbf{EBO} & \textbf{ReACT} & \textbf{DICE} & \textbf{DFM} \\
\hline
\textbf{Average} & 0.166 & 0.179 & 0.234 & 0.193 & 0.158 & 0.164 \\
\hline
\end{tabular}}
\label{tbl:overhead-analysis}
\end{table}
%%%%%%%%%%%%%%%%%%%%%%%%%%%%%%%%%%%%%%%%%%%%%%%%

\subsection{Overhead Analysis}
In addition to evaluating the performance of OOD detection methods, it is essential to analyze their computational overhead when deployed in real-world settings, such as edge computing environments. To achieve this goal, we compare the prediction time of OOD detection methods for each test sample. Table \ref{tbl:overhead-analysis} illustrates the per-sample inference overhead (in ms) for the six fastest out-of-distribution (OOD) detection methods. This result clearly demonstrates the advantages of DFM style methods. As one of the most accurate OOD detection methods, it also exhibits remarkable efficiency in making OOD predictions.  

% \subsection{Deep Insights}
% Something related to deep feature modeling
% TS-specific additions for better OOD detection

\section{Conclusion}
Out-of-distribution (OOD) detection is crucial for ensuring the reliability and robustness of machine learning models by identifying and mitigating risks posed by data deviating from the training distribution. Although extensively researched in the field of computer vision, these methods cannot be directly applied to the time-series domain. This paper aims to address this research gap by proposing a comprehensive evaluation pipeline for OOD detection in time-series data. 
%Our results showed that 1) most state-of-the-art OOD methods show poor performance on time-series data, and 2) deep feature modeling based OOD methods might be more beneficial for time-series OOD detection. 
The findings of our study reveal two key insights: First, most SOTA OOD detection methods perform poorly when applied to time-series data. Second, OOD detection approaches that leverage deep feature modeling appear to be more effective for time-series data, suggesting their potential for improving detection performance in this domain.   

\section{Acknowledgements}
This work has been funded in part by NSF, with award numbers \#1826967, \#1911095, \#2003279, \#2052809, \#2100237, \#2112167, \#2112665, and in part by PRISM and CoCoSys, centers in JUMP 2.0, an SRC program sponsored by DARPA.

\bibliography{aaai25}

\end{document}